\documentclass[runningheads]{llncs}

\usepackage[T1]{fontenc}
\usepackage{graphicx}
\usepackage{booktabs}
\usepackage{amsmath}
\usepackage{hyperref}
\usepackage{cleveref}

\begin{document}

\title{ArguAgent: AI-Supported Real-Time Grouping for Productive Argumentation in STEM Classrooms}

\titlerunning{AI-Supported Grouping for STEM Argumentation}

\author{Jennifer Kleiman\orcidID{0009-0000-3171-1954} \and
Yizhu Gao\orcidID{0000-0002-7791-3700} \and
Xin Xia \orcidID{0009-0009-1717-8511}\and
Zhaoji Wang \orcidID{0009-0007-2254-4438} \and
Zipei Zhu \orcidID{0009-0005-2886-0226} \and
Jongchan Park \orcidID{0000-0002-3257-125X}\and
Xiaoming Zhai\orcidID{0000-0003-4519-1931}\thanks{Corresponding author.}}

\authorrunning{J. Kleiman et al.}

\institute{AI4STEM Education Center,\\
 University of Georgia, Athens, GA, USA\\
\email{\{jennifer.kleiman, yizhu.gao,  xin.xia, zhaoji.wang, zipei.zhu, jongchan.park, xiaoming.zhai\}@uga.edu}}

\maketitle

\begin{abstract}
Argumentation is a core practice in STEM education, but its productivity depends on who participates and how they interact. Higher-achieving students often dominate the talk and decision-making, while lower-achieving peers may disengage, defer, or comply without contributing substantive reasoning. Forming groups strategically based on students’ stances and argumentation skills could help foster inclusive, evidence-based discourse. In practice, however, teachers are constrained in implementing this grouping strategy because it requires real-time insight into students’ positions and the quality of their argumentation, information that is difficult to assess reliably and at scale during instruction. We present a generative AI-powered system, ArguAgent, that creates groups optimizing for stance heterogeneity while constraining argumentation quality differences to $\pm$1 level on a validated learning progression. ArguAgent uses a two-component assessment pipeline: first scoring student arguments on a 0--4 rubric, then clustering positions via semantic analysis. We validated the scoring component against human expert consensus (Krippendorff's $\alpha$ = 0.817) using 200 expert-generated scores. Testing three OpenAI models (GPT-4o-mini, GPT-5.1, GPT-5.2) with identical calibrated prompts, we found that systematic prompt engineering informed by human disagreement analysis contributed 89\% of scoring improvement (QWK: 0.531 $\rightarrow$ 0.686), while model upgrades contributed an additional 11\% (QWK: 0.686 $\rightarrow$ 0.708). Simulation testing across 100 classes demonstrated that the grouping algorithm achieves 95.4\% of groups that meet both design criteria, a 3.2$\times$ improvement over random assignment. These results suggest ArguAgent can enable real-time, theoretically grounded grouping that promotes productive STEM argumentation in classrooms.

\end{abstract}

\section{Introduction}

Argumentation is widely recognized as a core practice in STEM education because it supports learners in constructing, evaluating, and revising evidence-based explanations and models \cite{berland2011classroom,osborne2010argument,ngss2013next}. However, productive argumentation is difficult to realize in typical classrooms \cite{berland2011classroom,osborne2010argument}. In small-group discussions, participation often becomes uneven: higher-achieving or more confident students may dominate talk and decision making, while lower-achieving peers may disengage, defer, or comply without contributing substantive reasoning \cite{webb1991,cohen1994,kumpulainen1999}. Such interaction patterns can limit opportunities for students to engage in key argumentative moves (e.g., coordinating claims with evidence and reasoning, and responding to counterarguments) \cite{toulmin1958,kuhn2003,osborne2016}.

Research suggests that group composition is a consequential design variable for argumentation outcomes \cite{webb1991,cohen1994}. In particular, groups benefit from genuine differences in positions because disagreement can elicit deliberative argumentation---engagement with alternative claims and evidence rather than superficial consensus \cite{iordanou2020,asterhan2016}. At the same time, large skill gaps can suppress equitable participation; narrower ability ranges can better support mutual regulation and sustained reasoning \cite{zillmer2018,webb1991}. Together, these findings motivate strategic grouping that simultaneously (a) promotes position heterogeneity to enable productive disagreement and (b) constrains within-group skill differences so that all members can participate meaningfully.

Despite its promise, strategic grouping is rarely implemented in day-to-day instruction because it requires timely, class-wide insight into what students think (their positions) and how well they argue (argumentation quality) \cite{dillenbourg1999,vanleeuwen2013}. Assessing argumentation quality reliably during instruction is particularly challenging, even with validated learning progressions and analytic rubrics \cite{osborne2016}. As a result, teachers often rely on ad hoc grouping heuristics that are not grounded in real-time evidence about student arguments \cite{cohen1994}.

To address this challenge, we present \textit{ArguAgent}, a generative AI--powered system that supports collaborative argumentation by linking automated argument assessment to group formation. Students first produce an individual written argument; ArguAgent then scores argument quality using a validated 0--4 learning progression \cite{osborne2016} and clusters student positions via semantic analysis. These estimates feed a grouping algorithm that maximizes position heterogeneity while constraining within-group quality gaps to $\pm$1 level. This paper focuses on validating the pre-discussion scoring and grouping components.

We address three research questions:
\begin{enumerate}
\item How accurately does ArguAgent score argumentation quality compared to trained human coders?
\item What contributes more to scoring accuracy: prompt calibration or model selection?
\item Does the grouping algorithm produce groups that satisfy the design criteria (position heterogeneity and bounded quality differences)?
\end{enumerate}

\section{Learning to Argue in Small Groups in STEM Classrooms}

Argumentation is not a single skill but a family of related competencies. Research distinguishes between constructing arguments and critiquing them~\cite{osborne2016}, between evaluating existing arguments and producing new ones~\cite{evagorou2023}, and between arguing for one's own position and engaging with opposing views~\cite{kuhn2003}. These sub-skills are correlated but distinct. A student who can identify weak evidence in another's argument may struggle to marshal evidence for their own.

Despite this complexity, research on argumentation development converges on a consistent finding: these competencies improve through sustained dialogic practice. Kuhn and Udell~\cite{kuhn2003} found that students who engaged in peer dialogues over 16 sessions showed increased use of counterargument strategies compared to a non-dialogue condition. Subsequent work has replicated this across diverse populations and age groups~\cite{iordanou2021,crowell2014,kuhn2024,iordanou2019}.

Not all peer interaction produces equivalent gains. Iordanou and Kuhn~\cite{iordanou2020} directly compared two conditions: students arguing with opposing-view peers versus same-side peers, with identical evidence. The opposing-view condition produced significantly greater gains, particularly in attending to opposing evidence and constructing arguments that weakened the other position. This finding aligns with the distinction between deliberative and disputative argumentation~\cite{asterhan2016}. Deliberative argumentation is issue-driven and characterized by a willingness to engage with alternative perspectives. This is distinct from disputative argumentation, where participants are arguing to win, or consensual argumentation, where the goal is to reach expedient agreement without necessarily critiquing alternative hypotheses.

Fostering deliberative argumentation is therefore desirable, but not guaranteed. Research has identified several factors of group dynamics that influence whether disagreement becomes deliberative. Of particular relevance to this work, Zillmer and Kuhn~\cite{zillmer2018} found that partners of similar ability can effectively regulate one another's thinking, alternating in serving metacognitive functions for each other. Relatedly, Webb's~\cite{webb1991} review of small-group learning in mathematics found that narrower ability ranges support more equal participation, while wide ability ranges can marginalize some members from productive interaction.

These findings lead us to two design principles for grouping students in argumentation activities: (1) Groups should contain members with different positions to enable deliberative argumentation, and (2) skill levels should be similar enough that all members can participate meaningfully and regulate each other's contributions.

To operationalize "similar skill level" in argumentation, we rely on Osborne et al.'s learning progression work~\cite{osborne2016}. Their research shows how argumentation development can be related to a hierarchical series of states: students first make claims, then learn to ground them in evidence, then to connect evidence to claims through reasoning, and finally to engage with opposing positions. Each transition represents a qualitatively different kind of argument, not a better version of the same one. Our $\pm$1 constraint is grounded in this developmental structure, where adjacent LP levels identify students whose argumentation differs enough to create productive tension but not so much that the gap prevents mutual engagement.

To implement this in the classroom, teachers would need to assess both what students think and how well they argue. However, applying argumentation learning progression frameworks manually is time-intensive and difficult even for researchers.

\section{Automated Scoring of Scientific Argumentation}

Recent advances in automated assessment suggest a path forward. Machine learning approaches have achieved reliable scoring of scientific arguments across a range of tasks and rubrics~\cite{wilson2024,zhai2023,zhai2020}, and large language models with prompt engineering strategies including few-shot and chain-of-thought have shown strong performance on argument evaluation tasks~\cite{lee2024}. These systems have been developed primarily for summative assessment and have not been applied to the grouping problem. Wilson et al.~\cite{wilson2024} demonstrated that analytic scoring of argumentation components outperforms holistic approaches when aligned to the Osborne et al.~\cite{osborne2016} learning progression, a finding directly relevant to our work.

Despite these advances, two gaps remain. First, few studies have applied LLMs to learning progression frameworks that capture the full developmental trajectory from simple claims through coordinated arguments with rebuttals. Second, automated assessment has been developed primarily for summative purposes. The question of whether these tools can support formative instructional decisions in real time has received little attention. Recent work has developed computer-supported collaborative argumentation environments~\cite{song2024insights}, but no prior work has combined automated stance detection with argumentation quality assessment to form instructionally optimal groups. ArguAgent represents an attempt to address this problem.

\section{Methods}

\subsection{ArguAgent System Overview}

ArguAgent is designed to facilitate small-group STEM argumentation by addressing the grouping problem, helping teachers form discussion groups of 2--4 students that foster productive argumentation while minimizing ability gaps that might inhibit participation. In this system, students individually learn about a science phenomenon and are prompted to make a claim about it citing evidence and reasoning. The system analyzes their responses in two ways, using a 0--4 quality scale based on Osborne, and creating 2-4 stance clusters (e.g., ``all objects deform,'' ``only some objects deform''), then suggests grouping to the teacher who can adjust then launch discussions in a CSCL environment with real-time analysis and scaffolding.

\subsection{Dataset} 
We generated 200 simulated student responses to a science argumentation task about object deformation during collisions. The prompt asked students to respond to the claim that ``all objects change shape when they collide.'' Five researchers collaboratively authored responses designed to span the full range of argumentation quality (Levels 0--4) with realistic variation in language, complexity, and scientific accuracy, drawing on our experience with middle school student writing. Responses ranged from 1--4 sentences. \Cref{tab:examples} shows representative examples at each rubric level; all examples shown received unanimous coder agreement.

\begin{table}[t]
\caption{Example Responses at Each Argumentation Quality Level}\label{tab:examples}
\centering
\small
\begin{tabular}{@{}cp{9.5cm}@{}}
\toprule
Level & Example Response \\
\midrule
0 & ``Video B was my favorite one.'' \\
1 & ``I think objects never change shape unless they break.'' \\
2 & ``Yes they are right because the tennis ball got flat and the water balloon squished.'' \\
3 & ``I agree because when things hit, force happens. The water balloon changed shape in the video C so the bowling ball must change too but maybe it's too little to see.'' \\
4 & ``No, that's too extreme. All objects technically deform even a little bit because of Newton's third law. But we're talking visible change here. The bowling ball and cars look the same after. The squishy stuff like the balloon deform a lot.'' \\
\bottomrule
\end{tabular}
\end{table}

The dataset was divided into two sets: a \textit{Calibration Set} (rows 1--21, $n$=21) used for human coder training and consensus building, excluded from AI-human comparison metrics to avoid data leakage; and a \textit{Test Set} (rows 22--200, $n$=179) used for AI-human comparison and reliability analysis.

\subsection{Argumentation Quality Coding}

To score argumentation quality, we adapted the learning progression developed and validated by Osborne et al.~\cite{osborne2016}, which captures increasingly sophisticated coordination of claims, evidence, reasoning, and rebuttals. The original framework distinguishes between construction and critique skills; we focus on construction for the initial response assessment. \Cref{tab:rubric} presents the adapted rubric.

\begin{table}[t]
\caption{ArguAgent Argumentation Quality Rubric}\label{tab:rubric}
\centering
\small
\begin{tabular}{@{}clp{7.5cm}@{}}
\toprule
Level & Label & Criteria \\
\midrule
0 & No Response & No claim, evidence, or reasoning. Includes ``I don't know,'' blank responses, or irrelevant content. \\
1 & Claim Only & A relevant claim without supporting evidence. Note: Elaboration on a claim without citing evidence remains Level 1. \\
2 & Claim + Evidence & A claim backed by observations, data, or examples. Evidence must be explicitly cited, not merely implied. \\
3 & Argument & Claim + evidence + reasoning that explains why the evidence supports the claim. All three components must be present and connected. \\
4 & Complete Argument & A Level 3 argument that additionally addresses counterarguments or considers alternate positions. \\
\bottomrule
\end{tabular}
\end{table}


We employed four coders to score students responses: two graduate research assistants with science education backgrounds, one post-doc, and one faculty researcher. All received training on the argumentation rubric prior to coding. All four coders independently scored responses 1--21, recording both the argumentation level (0--4) and a summary of the student's main claim. Following independent coding, the team convened for a consensus meeting to discuss all disagreements. 

We identified five critical scoring principles that helped resolve ambiguity:
\begin{enumerate}
\item \textbf{Elaboration will not count as reasoning.} Expanding on a claim without citing specific evidence will not count as reasoning, just making more claims, so remains Level 1.
\item \textbf{Evaluate only explicit content.} Scorers should not infer evidence or reasoning the student did not state.
\item \textbf{Logical chains are not evidence.} Deductive reasoning from assumptions does not count as cited evidence.
\item \textbf{Reasoning is not restating.} Simply saying ``because [evidence]'' is not reasoning; the response must explain why the evidence supports the claim.
\item \textbf{Mechanistic explanations count as reasoning.} Physical or conceptual explanations of how evidence supports a claim qualify as reasoning (when evidence is present).
\end{enumerate}

These principles were incorporated into both the human codebook and the AI scoring prompt.

Four coders then independently scored responses 22--48 ($n$=27). Inter-rater reliability was computed using Krippendorff's alpha for ordinal data, which handles multiple coders and is appropriately conservative for ordinal scales~\cite{hayes2007}.  As shown in \Cref{tab:irr}, the alpha of 0.817 exceeds the 0.80 threshold recommended for reliable data~\cite{krippendorff2004}.

\begin{table}[t]
\caption{Inter-Rater Reliability Results}\label{tab:irr}
\centering
\begin{tabular}{@{}lr@{}}
\toprule
Metric & Value \\
\midrule
Krippendorff's Alpha (ordinal) & 0.817 \\
Percent Agreement (Exact) & 74.2\% \\
Percent Agreement ($\pm$1 level) & 90.6\% \\
\bottomrule
\end{tabular}
\end{table}

The remaining 152 responses (rows 49--200) were distributed among coder pairs. Each response was coded by two coders working independently. The consensus score was calculated as the average of the two coder scores, rounded to the nearest integer.

\subsubsection{AI Scoring and Calibration}

The AI implementation uses a large language model to analyze each students' response against the rubric. The prompt includes the rubric definitions, evidence evaluation criteria, the five scoring principles, and a decision tree for systematic rubric application (e.g., ``Does the response contain a relevant claim? If no $\rightarrow$ Level 0. If yes, does it cite specific evidence? If no $\rightarrow$ Level 1\ldots''). The AI model returns the assigned level (0--4), an explanation of the score, the extracted claim, and highlights identifying claim, evidence, reasoning, and rebuttal components in the text. The full prompt is available in the supplementary materials.\footnote{Scoring prompts, simulation code, and data are available at \url{https://github.com/jenniferbk/arguagent-aied-2026}.}

We systematically compared three generations of models (GPT-4o-mini, GPT-5.1, GPT-5.2) while iteratively refining the scoring prompt based on disagreement analysis. We report four agreement metrics: Quadratic Weighted Kappa (QWK), which measures ordinal agreement adjusted for chance, weighting larger disagreements more heavily~\cite{cohen1968}; exact match (percentage of identical scores); mean absolute error (MAE, average magnitude of score differences); and Pearson correlation.

\textbf{Baseline (GPT-4o-mini, Uncalibrated Prompt).} Initial scoring used GPT-4o-mini with a basic prompt containing only the rubric definitions. This version revealed significant systematic bias: Bias (AI $-$ Human) = $+$0.85 levels; Exact Match = 42.4\%; QWK = 0.531.

\textbf{Root Cause Analysis.} We identified all AI-human disagreements in the baseline and identified t three primary error patterns in order of frequency: (1) \textit{Elaboration scored as reasoning} (most common): AI credited claim elaborations as reasoning, inflating Level 1 responses to Level 2 or 3; (2) \textit{Reasoning without evidence}: AI accepted logical statements without grounded observations as sufficient for Level 3; (3) \textit{Inferred components}: AI filled in logical gaps students did not explicitly state, crediting implied rather than stated evidence or reasoning. Together, these three patterns accounted for the majority of baseline disagreements.

\textbf{Calibrated Prompt Development.} Based on this analysis, we developed a calibrated incorporating: (1) evidence evaluation criteria distinguishing observations from elaborations, (2) the five scoring principles derived from human consensus norms, and (3) a scoring decision tree for systematic rubric application. To isolate the effects of prompt engineering from model capability, we ran all three models with the identical calibrated prompt on the test set ($n$=179).

\subsection{Student Stance Detection}

We conducted preliminary validation by comparing AI-extracted stance categories against human coder judgments. One coder independently classified all 200 responses into three stance categories: ``All objects change shape'' (ALL), ``Only some/no objects change shape'' (SOME/NO), or ``Unsure.'' AI-extracted claims from the earlier scoring were then programmatically classified into the same categories based on linguistic markers.

We initially explored embedding-based clustering using OpenAI's text-embedding-3-small model with cosine similarity thresholds, but found that embeddings fail to distinguish semantic opposites. For example, ``Objects DO change shape'' and ``Objects DON'T change shape'' have high embedding similarity (both discuss objects and shape change) despite representing opposite positions. The LLM-based approach handles these distinctions more reliably.

\subsection{Grouping Algorithm Simulation}

The grouping process has two stages. First, ArguAgent uses GPT-5.1 to process all student responses in a class and identify 2--4 natural \textit{position clusters} based on semantic similarity and stance (e.g., ``all objects deform,'' ``only soft objects deform,'' ``deformation depends on speed''). Each student is assigned to one cluster. Second, the grouping algorithm forms \textit{discussion groups} of 3 students, drawing from different position clusters while constraining argumentation quality differences.

\textbf{Design Criteria.} Groups were evaluated against two criteria: all student pairs within $\pm$1 level of argumentation quality, and the group contains multiple position clusters (stance heterogeneity).

\textbf{Why not a simple assignment algorithm?} Once students are classified by stance and level, a deterministic algorithm could in principle form valid groups. We use the LLM for the position clustering stage because stance categories are not predetermined, but emerge from the content of student responses and vary by topic. The subsequent group assignment uses a deterministic scoring-based optimization algorithm.

To evaluate the grouping algorithm, we ran a Monte Carlo simulation with 100 simulated classes of 24 students each. Level distributions matched the human-scored data (Level 0: 11\%, Level 1: 28\%, Level 2: 32\%, Level 3: 16\%, Level 4: 12\%)---proportions derived from our coded dataset. Each student was randomly assigned to one of four position clusters. The algorithm formed groups of 3 students per class, producing 800 groups across all simulations.

\subsubsection{Group Score Indicator.}

The algorithm optimizes a group score computed as:
\[
\text{GroupScore} = \text{PositionScore} + \text{LevelScore}
\]

\textbf{LevelScore} enforces the $\pm$1 level constraint as the primary objective: $-100$ if level difference $> 1$ (hard constraint violation); $+30$ if level difference $= 1$ (optimal zone of proximal development); $+10$ if level difference $= 0$ (acceptable, no scaffolding opportunity).

\textbf{PositionScore} rewards stance heterogeneity as a secondary objective: $+40$ if the group contains mixed positions; $-20$ if all members hold the same position.

Target thresholds in \Cref{tab:results} were established prior to analysis based on two considerations: the QWK $>$ 0.65 target aligns with the ``substantial agreement'' threshold on standard interpretation scales~\cite{landis1977}, the exact match $>$ 55\% target reflects a practical minimum for the $\pm$1 grouping constraint (since within-$\pm$1 agreement $>$ 80\% means most scoring errors will not affect group validity), and bias $< \pm$0.2 ensures no systematic directional error.

\section{Results}

\subsection{Argumentation Quality Scoring.}

\Cref{tab:results} presents scoring performance across all three models with the calibrated prompt. All models exceeded target thresholds. The best-performing configuration (GPT-5.2) achieved QWK = 0.708, indicating ``substantial agreement'' on standard interpretation scales~\cite{landis1977}.

\begin{table}[t]
\caption{Model Comparison with Calibrated Prompt}\label{tab:results}
\centering
\small
\begin{tabular}{@{}lcccc@{}}
\toprule
Metric & GPT-4o-mini & GPT-5.1 & GPT-5.2 & Target \\
\midrule
Exact Match & 50.3\% & 67.6\% & \textbf{68.7\%} & $>$55\% \\
Within $\pm$1 Level & \textbf{91.1\%} & 89.9\% & 89.9\% & $>$80\% \\
Mean Absolute Error & 0.60 & 0.45 & \textbf{0.44} & $<$0.6 \\
Bias (AI $-$ Human) & $-$0.11 & $-$0.06 & \textbf{$-$0.03} & $<\pm$0.2 \\
Pearson Correlation & 0.690 & 0.693 & \textbf{0.710} & $>$0.65 \\
QWK & 0.686 & 0.687 & \textbf{0.708} & $>$0.65 \\
\bottomrule
\end{tabular}
\end{table}

\Cref{tab:distribution} shows that GPT-5.2 produced a Level 4 count ($n$=22) matching the human distribution, while GPT-5.1 under-detected Level 4 ($n$=13). However, matching the distribution count does not imply accurate identification of the same responses (see error analysis below).

\begin{table}[t]
\caption{Level Distribution Comparison}\label{tab:distribution}
\centering
\begin{tabular}{@{}lccccc@{}}
\toprule
Level & Human ($n$) & 4o-mini & GPT-5.1 & GPT-5.2 \\
\midrule
0 & 20 & 16 & 13 & 12 \\
1 & 51 & 67 & 60 & 63 \\
2 & 57 & 51 & 62 & 62 \\
3 & 29 & 28 & 31 & 20 \\
4 & 22 & 17 & 13 & \textbf{22} \\
\bottomrule
\end{tabular}
\end{table}

\Cref{tab:decomposition} decomposes the sources of improvement. Prompt engineering contributed 89\% of the total scoring improvement, while model upgrades contributed 11\%.

\begin{table}[t]
\caption{Decomposing Improvement Sources}\label{tab:decomposition}
\centering
\begin{tabular}{@{}lcc@{}}
\toprule
Comparison & QWK Change & \% of Total \\
\midrule
Prompt engineering (uncalibrated $\rightarrow$ calibrated) & +0.155 & \textbf{89\%} \\
Model upgrade (4o-mini $\rightarrow$ 5.2, calibrated) & +0.022 & \textbf{11\%} \\
\textbf{Total improvement} & \textbf{+0.177} & \textbf{100\%} \\
\bottomrule
\end{tabular}
\end{table}

This decomposition reveals that systematic prompt engineering informed by human disagreement analysis contributed 89\% of the total scoring improvement, while model upgrades contributed an additional 11\%.

\textbf{Practical Implication:} Schools with budget constraints can achieve near-optimal scoring accuracy using GPT-4o-mini with the calibrated prompt, at substantially lower cost than reasoning-focused models.

\subsubsection{Error Analysis}

Analysis of remaining disagreements in the best-performing configuration (GPT-5.2) revealed systematic patterns. Of 179 test responses, 56 (31.3\%) showed disagreement with human consensus.

\textbf{Off by 1 Level} ($n$=39, 21.8\% of test set): These disagreements primarily involved borderline cases where the response showed partial evidence of the higher level. Common patterns included: Level 1 vs.\ 2 (ambiguity about whether cited information constituted ``evidence'' or ``elaboration''); Level 2 vs.\ 3 (unclear whether explanatory text constituted ``reasoning'' or merely restated evidence); Level 3 vs.\ 4 (disagreement about whether acknowledgment of complexity constituted ``considering alternate positions'').

\textbf{Off by 2+ Levels} ($n$=17, 9.5\% of test set): Larger disagreements concentrated primarily at Level 0, where AI sometimes identified a claim where human coders judged the response as non-responsive or irrelevant.

\textbf{Level 4 Detection Analysis.} Level 4 responses, which require rebuttals or consideration of alternate positions, proved most challenging. GPT-5.1 correctly identified 10 of 22 human-scored Level 4 responses (45\% recall) with 3 false positives. GPT-5.2 improved to 14 of 22 (64\% recall) but with 9 false positives. The matching distribution count (GPT-5.2 assigned 22 Level 4 scores) occurred because 8 misses were balanced by 8 false positives.

Analysis of Level 4 errors reveals two systematic patterns. \textit{False negatives} occurred when students addressed counterarguments implicitly rather than explicitly. For example, one student wrote: ``We cannot see with our eye because it is very fast and very small.'' Human coders recognized this as addressing the counterargument that no deformation was observed, but the AI scored it as Level 1 (claim only). The AI may require explicit linguistic markers such as ``however'' or ``on the other hand'' to recognize rebuttal structures. Upon reflection, the researchers agreed that the AI assessment may more accurately reflect the rubric as written.

\textit{False positives} occurred when the AI over-credited conditional language as rebuttals. Statements like ``it depends on the definition'' or ``I don't agree fully'' were scored as Level 4 by the AI but Level 2--3 by humans, who interpreted these as qualifying the claim rather than genuinely engaging with alternate positions.

This boundary between ``qualifying a claim'' and ``engaging with counterarguments'' is inherently difficult to operationalize. When any human coder assigned Level 4, coders agreed unanimously only 37\% of the time, with 63\% of Level 4 assignments disputed. Level 3 vs.\ 4 disagreements accounted for 40\% of all human scoring disputes (19 of 47 disagreements). The AI's 64\% recall rate, while imperfect, operates on a task where trained human coders frequently disagree.

\subsection{Stance Detection Validation}

The AI achieved 94\% accuracy on SOME/NO-stance responses but only 74\% on ALL-stance responses. The primary error pattern was misclassifying ``ALL'' responses as ``SOME'' when students used hedging language (e.g., ``I think all objects probably change\ldots''). Cohen's Kappa of 0.690 indicates substantial agreement, though below the argumentation quality scoring performance. This validation is preliminary, relying on a single coder; future work will establish multi-coder reliability for stance classification.

\subsection{Grouping Algorithm Simulation}

\Cref{tab:simulation} presents the simulation results. The algorithm achieved 95.4\% of groups meeting both design criteria simultaneously, a 3.2$\times$ improvement over random assignment (30.3\%).

\begin{table}[t]
\caption{Grouping Algorithm Simulation Results}\label{tab:simulation}
\centering
\begin{tabular}{@{}lcccc@{}}
\toprule
Algorithm & $\pm$1 Level & Mixed Positions & Both Criteria & vs.\ Random \\
\midrule
Random assignment & $\sim$35\% & $\sim$75\% & 30.3\% & 1.0$\times$ \\
ArguAgent grouping & \textbf{96.8\%} & \textbf{98.6\%} & \textbf{95.4\%} & \textbf{3.2$\times$} \\
\bottomrule
\end{tabular}
\end{table}

\section{Discussion}

\subsection{Design Contributions}

ArguAgent addresses the need to form small discussion groups that balance stance heterogeneity with quality homogeneity for collaborative argumentation, grounded in research on productive disagreement~\cite{iordanou2020} and peer regulation among similar-ability partners~\cite{zillmer2018}.

\subsection{Primacy of Prompt Engineering Over Frontier Models}

Our most striking finding is that prompt engineering contributed 89\% of scoring improvement, with model upgrades contributing only 11\%. The calibration process that yielded the five scoring principles is the primary contribution, not the choice of model. These principles encode tacit knowledge that distinguishes expert from naive rubric application. They are transferable: the same calibrated prompt works across three model generations with minimal performance variation between GPT-4o-mini (QWK = 0.686) and GPT-5.1 (QWK = 0.687). Schools can thus achieve reliable scoring without access to expensive frontier models.

That said, model choice does matter for specific capabilities. GPT-5.2 showed improved Level 4 recall (64\% vs.\ 45\% for GPT-5.1), though both models struggle with this highest level, as did human coders. Applications requiring precise identification of sophisticated arguments may need additional calibration or human oversight.

\subsection{Implications for Practice}

If validated in classroom settings, ArguAgent could enable teachers to employ principled grouping without the manual assessment burden that currently limits practice. Research suggests that structured discussion groups improve argumentation outcomes, yet surveys indicate fewer than a third of teachers routinely use them~\cite{buchs2017}. Automated assessment could help address this barrier, as the system handles assessment and grouping for a class of 30 students in under two minutes. The system is designed to support teacher agency: all scores include explanations, and teachers can review and override both scores and groupings before launching discussion.

\subsection{Limitations and Ethical Considerations}

Our grouping algorithm is theoretically motivated but untested in practice. All 200 responses addressed a single topic (object deformation) and were researcher-generated. While constructed to reflect realistic variation in middle school writing, they may not capture the full range of student language, misconceptions, and response patterns. Additionally, we validated scoring reliability and not whether ArguAgent-formed groups improve learning outcomes. 

Stance detection validation (85\% agreement, $\kappa$ = 0.690) is preliminary, relying on a single coder. In deployment, stance detection errors, particularly the 26\% misclassification rate for ``ALL'' stances, could produce groups that fail the heterogeneity criterion despite the algorithm functioning correctly.

The system currently depends on commercial LLM APIs with cost, latency, and data privacy considerations. The 0--4 rubric captures argument construction but not critique, the other major skill in the Osborne et al.~\cite{osborne2016} framework. Bias and fairness remain ongoing concerns, particularly given prior findings of differential scoring for English learner students~\cite{wilson2024}. Future work must examine performance across diverse student populations.

ArguAgent is intended for formative use to support instructional decision-making rather than summative assessment. We caution against summative applications without substantially stronger validation and human oversight, and we emphasize the need to clearly communicate system limitations to users and stakeholders.

\subsection{Future Work}

We plan to pilot the system in Spring 2026 with both teachers and students. Our classroom pilot will gather data allowing us to more rigorously validate the argumentation analysis and stance detection with real student data, and understand how students and teachers engage in small group discussions aided by ArguAgent. We will also test the system across multiple STEM topics. Similarly, testing whether the five scoring principles transfer across topics will determine how much domain-specific tuning is required for deployment. Additional directions include extending the system to capture critique and exploring the use of locally hosted open source models to reduce privacy and security concerns, and testing across diverse model families beyond OpenAI.

\section{Conclusion}

We presented ArguAgent, a system combining automated quality assessment with position clustering to form groups optimized for productive argumentation. Validation shows that LLM-based scoring achieves substantial agreement with human consensus (QWK = 0.708) after calibration, with prompt engineering contributing 89\% of scoring improvement versus model upgrades contributing 11\%. Preliminary validation of position clustering showed 85\% agreement with human stance judgments ($\kappa$ = 0.690). Simulation testing demonstrated that the grouping algorithm achieves 95.4\% of groups meeting both design criteria, a 3.2$\times$ improvement over random assignment.

The next step is classroom validation. If ArguAgent-grouped discussions produce better argumentation outcomes than random grouping, the system could help teachers implement evidence-based grouping practices at scale, potentially increasing the adoption of structured argumentation activities in science classrooms.

\begin{credits}

\subsubsection{\ackname} This work was supported by the Institute of Education Sciences, U.S. Department of Education, through Grant No.\ R305C240010 (GENIUS). Any opinions, findings, conclusions, or recommendations expressed in this material are those of the authors and do not necessarily reflect the views of the Institute of Education Sciences.

\subsubsection{\discintname} The authors have no competing interests to declare that are relevant to the content of this article.
\end{credits}

\bibliographystyle{splncs04}
\bibliography{reference}

@book{ngss2013next,
  title={Next generation science standards: For states, by states},
  author={NGSS Lead States},
  year={2013},
  publisher={National Academies Press}
}

@article{osborne2010argument,
  title={An argument for arguments in science classes},
  author={Osborne, Jonathan Francis},
  journal={Phi Delta Kappan},
  volume={91},
  number={4},
  pages={62--65},
  year={2010},
  publisher={SAGE Publications Sage CA: Los Angeles, CA}
}

@article{berland2011classroom,
  title={Classroom communities' adaptations of the practice of scientific argumentation},
  author={Berland, Leema K and Reiser, Brian J},
  journal={Science Education},
  volume={95},
  number={2},
  pages={191--216},
  year={2011},
  publisher={Wiley Online Library}
}

@article{cohen1994,
  title={Restructuring the classroom: Conditions for productive small groups},
  author={Cohen, Elizabeth G},
  journal={Review of Educational Research},
  volume={64},
  number={1},
  pages={1--35},
  year={1994},
  publisher={SAGE}
}

@article{kumpulainen1999,
  title={The situated dynamics of peer group interaction: An introduction to an analytic framework},
  author={Kumpulainen, Kristiina and Mutanen, Mika},
  journal={Learning and Instruction},
  volume={9},
  pages={449--473},
  year={1999},
  publisher={Elsevier}
}

@book{dillenbourg1999,
  title={Collaborative Learning: Cognitive and Computational Approaches},
  editor={Dillenbourg, Pierre},
  year={1999},
  publisher={Elsevier Science}
}

@article{vanleeuwen2013,
  title={Teacher interventions in a synchronous, co-located CSCL setting: Analyzing focus, means, and temporality},
  author={Van Leeuwen, Anouschka and Janssen, Jeroen and Erkens, Gijsbert and Brekelmans, Mieke},
  journal={Computers in Human Behavior},
  volume={29},
  number={3},
  pages={1377--1386},
  year={2013},
  publisher={Elsevier}
}

@book{toulmin1958,
  title={The Uses of Argument},
  author={Toulmin, Stephen E},
  year={1958},
  publisher={Cambridge University Press}
}

@article{osborne2016,
  title={The development and validation of a learning progression for argumentation in science},
  author={Osborne, Jonathan F and Henderson, J Bryan and MacPherson, Anna and Szu, Emily and Wild, Andrew and Yao, Shi-Ying},
  journal={Journal of Research in Science Teaching},
  volume={53},
  pages={821--846},
  year={2016},
  publisher={Wiley}
}

@article{evagorou2023,
  title={What do we really know about students' written arguments? Evaluating written argumentation skills},
  author={Evagorou, Maria and Papanastasiou, Elena and Vrikki, Maria},
  journal={European Journal of Science and Mathematics Education},
  volume={11},
  pages={615--634},
  year={2023}
}

@article{kuhn2003,
  title={The Development of Argument Skills},
  author={Kuhn, Deanna and Udell, Wadiya},
  journal={Child Development},
  volume={74},
  pages={1245--1260},
  year={2003},
  publisher={Wiley}
}

@article{iordanou2021,
  title={``Argue With Me'': A Method for Developing Argument Skills},
  author={Iordanou, Kalypso and Rapanta, Chrysi},
  journal={Frontiers in Psychology},
  volume={12},
  pages={631203},
  year={2021}
}

@article{crowell2014,
  title={Developing Dialogic Argumentation Skills: A 3-year Intervention Study},
  author={Crowell, Amanda and Kuhn, Deanna},
  journal={Journal of Cognition and Development},
  volume={15},
  pages={363--381},
  year={2014},
  publisher={Taylor \& Francis}
}

@article{kuhn2024,
  title={Enriching Thinking Through Discourse},
  author={Kuhn, Deanna and Bruun, Samantha and Geithner, Christopher},
  journal={Cognitive Science},
  volume={48},
  pages={1--17},
  year={2024},
  publisher={Wiley}
}

@article{iordanou2019,
  title={Learning by arguing},
  author={Iordanou, Kalypso and Kuhn, Deanna and Matos, Florencia and Shi, Yuchen and Hemberger, Laura},
  journal={Learning and Instruction},
  volume={63},
  pages={101207},
  year={2019},
  publisher={Elsevier}
}

@article{iordanou2020,
  title={Contemplating the Opposition: Does a Personal Touch Matter?},
  author={Iordanou, Kalypso and Kuhn, Deanna},
  journal={Discourse Processes},
  volume={57},
  pages={343--359},
  year={2020},
  publisher={Taylor \& Francis}
}

@article{asterhan2016,
  title={Argumentation for Learning: Well-Trodden Paths and Unexplored Territories},
  author={Asterhan, Christa S C and Schwarz, Baruch B},
  journal={Educational Psychologist},
  volume={51},
  pages={164--187},
  year={2016},
  publisher={Taylor \& Francis}
}

@article{zillmer2018,
  title={Do similar-ability peers regulate one another in a collaborative discourse activity?},
  author={Zillmer, Nicole and Kuhn, Deanna},
  journal={Cognitive Development},
  volume={45},
  pages={68--76},
  year={2018},
  publisher={Elsevier}
}

@article{webb1991,
  title={Task-Related Verbal Interaction and Mathematics Learning in Small Groups},
  author={Webb, Noreen},
  journal={Journal for Research in Mathematics Education},
  volume={22},
  pages={366--389},
  year={1991},
  publisher={NCTM}
}

@article{wilson2024,
  title={Using automated analysis to assess middle school students' competence with scientific argumentation},
  author={Wilson, Christopher D and others},
  journal={Journal of Research in Science Teaching},
  volume={61},
  pages={38--69},
  year={2024},
  publisher={Wiley}
}

@article{zhai2023,
  title={Assessing Argumentation Using Machine Learning and Cognitive Diagnostic Modeling},
  author={Zhai, Xiaoming and Haudek, Kevin C and Ma, Wenchao},
  journal={Research in Science Education},
  volume={53},
  pages={405--424},
  year={2023},
  publisher={Springer}
}

@article{zhai2020,
  title={Applying machine learning in science assessment: a systematic review},
  author={Zhai, Xiaoming and Yin, Yue and Pellegrino, James W and Haudek, Kevin C and Shi, Li},
  journal={Studies in Science Education},
  volume={56},
  pages={111--151},
  year={2020},
  publisher={Taylor \& Francis}
}

@article{lee2024,
  title={Applying large language models and chain-of-thought for automatic scoring},
  author={Lee, Gyeong-Geon and Latif, Ehsan and Wu, Xiaoming and Liu, Ningyu and Zhai, Xiaoming},
  journal={Computers and Education: Artificial Intelligence},
  volume={6},
  pages={100213},
  year={2024},
  publisher={Elsevier}
}

@article{song2024insights,
  title={Insights into critical discussion: Designing a computer-supported collaborative space for middle schoolers},
  author={Song, Yi and Ferretti, Ralph P and Sabatini, John and Cui, Wenting},
  journal={ETS Research Report Series},
  volume={2024},
  pages={1--20},
  year={2024},
  publisher={Wiley}
}

@article{hayes2007,
  title={Answering the Call for a Standard Reliability Measure for Coding Data},
  author={Hayes, Andrew F and Krippendorff, Klaus},
  journal={Communication Methods and Measures},
  volume={1},
  pages={77--89},
  year={2007},
  publisher={Taylor \& Francis}
}

@book{krippendorff2004,
  title={Content Analysis: An Introduction to Its Methodology},
  author={Krippendorff, Klaus},
  edition={2nd},
  year={2004},
  publisher={Sage Publications},
  address={Thousand Oaks}
}

@article{buchs2017,
  title={Challenges for cooperative learning implementation: Reports from elementary school teachers},
  author={Buchs, C{\'e}line and Filippou, Dimitra and Pulfrey, Caroline and Volp{\'e}, Yann},
  journal={Journal of Education for Teaching},
  volume={43},
  number={3},
  pages={296--306},
  year={2017},
  doi={10.1080/02607476.2017.1321673}
}

@article{cohen1968,
  title={Weighted kappa: Nominal scale agreement provision for scaled disagreement or partial credit},
  author={Cohen, Jacob},
  journal={Psychological Bulletin},
  volume={70},
  number={4},
  pages={213--220},
  year={1968},
  doi={10.1037/h0026256}
}

@article{landis1977,
  title={The measurement of observer agreement for categorical data},
  author={Landis, J. Richard and Koch, Gary G.},
  journal={Biometrics},
  volume={33},
  number={1},
  pages={159--174},
  year={1977}
}

\end{document}